\documentclass[pagenumber]{apiems}
\conferenceyear{2026}

\usepackage{amsmath,amsfonts,bm}

\def\eqref#1{equation~\ref{#1}}

\def\1{\bm{1}}

\def\vh{{\bm{h}}}

\def\vl{{\bm{l}}}

\def\vo{{\bm{o}}}

\def\vu{{\bm{u}}}

\def\vw{{\bm{w}}}

\def\mA{{\bm{A}}}
\def\mB{{\bm{B}}}
\def\mC{{\bm{C}}}

\def\mM{{\bm{M}}}

\def\mX{{\bm{X}}}

\DeclareMathAlphabet{\mathsfit}{\encodingdefault}{\sfdefault}{m}{sl}
\SetMathAlphabet{\mathsfit}{bold}{\encodingdefault}{\sfdefault}{bx}{n}

\usepackage{enumitem}
\usepackage{microtype}
\usepackage{pifont}
\usepackage{xcolor}
\usepackage{subcaption}
\makeatletter
\patchcmd{\ttl@straight@i}{\par\nobreak}{\par}{}{}
\patchcmd{\@afterheading}{\clubpenalty \@M}{\clubpenalty\z@}{}{}
\makeatother
\newcommand{\modelname}{MASCIT}
\newcommand{\cmark}{{\color[HTML]{0C6B37}\ding{51}}}
\newcommand{\xmark}{{\color[HTML]{BC2023}\ding{55}}}

\graphicspath{{figures/}}

\title{MASCIT: A Mask-Aware State Space Classifier\\for Naturally Irregular Time Series}

\author{%
  \authorblock{Yoo-Min Jung}{%
    Department of Industrial Engineering\\
    Seoul National University, Seoul, Republic of Korea\\
    Tel: (+82) 2-880-7361, Email: \email{pamela7384@gmail.com}}
  \authorblock{Hyeon-Gi Kim}{%
    Department of Industrial Engineering\\
    Seoul National University, Seoul, Republic of Korea\\
    Tel: (+82) 2-880-7361, Email: \email{kimhun0505@snu.ac.kr}}
  \authorblock{Jonghun Park}{%
    Department of Industrial Engineering\\
    Seoul National University, Seoul, Republic of Korea\\
    Tel: (+82) 2-880-7361, Email: \email{jonghun@snu.ac.kr}}
}
\runninghead{Jung, Kim \& Park}

\begin{document}
\twocolumn[{%
  \maketitle
  \begin{APIEMSabstract}
  Naturally irregular time series combine asynchronous observations, missing values, unequal lengths, and nonuniform sampling, while dense adapters can discard temporal structure.
  We propose a mask-aware state space classifier for irregular time series (MASCIT), which supplies observation masks to the encoder and excludes invalid steps from gated temporal aggregation.
  Across 34 irregular time series datasets, MASCIT yielded the strongest aggregate point estimate and was the only evaluated neural model with three-seed results on every dataset.
  MASCIT retained the lowest point rank across six overlapping irregularity indicators, while factorial ablations favored partial over full selectivity.
  These results support selective state space models as effective, executable backbones for naturally irregular time series classification.
  \end{APIEMSabstract}
  \APIEMSkeywords{irregular time series, time series classification, Mamba, observation masks, partial selectivity}
  \vspace{6pt}
}]
\makeauthorthanks

\section{Introduction}
\label{sec:introduction}

Event-driven clinical, mobility, and wearable time series combine asynchronous channels, missing observations, unequal lengths, and nonuniform sampling intervals~\citep{spinnato2026pyrregular}.
Their classification depends on both the sequence model and the temporal information retained by a learnable representation.

Structured state space models (SSMs) combine recurrent state updates with parallelizable training, providing an efficient backbone for long-sequence modeling~\citep{gu2021lssl,gu2022s4}.
Mamba introduces input-conditioned selection, allowing the recurrence to retain or discard information according to sequence content while preserving linear sequence complexity~\citep{gu2023mamba}.
This general-purpose mechanism does not itself represent irregular observations, nor does time series classification (TSC) necessarily benefit from applying selectivity uniformly throughout the state space layer.
Partial selectivity instead treats input dependence as an architectural choice, allowing adaptive and invariant dynamics to coexist.
Our previous work adapted Mamba to multivariate classification with one partially selective layer and gated temporal aggregation~\citep{jung2026mambasl}.
That work used regularly sampled archives, so its value-only input hid observation patterns and its gate normalization retained padding.
Transfer to naturally irregular data therefore requires mask-aware encoder and readout interfaces, consistent with missingness-aware recurrent classifiers~\citep{che2018recurrent,cao2018brits}.

PYRREGULAR provides a shared representation and protocol for this evaluation~\citep{spinnato2026pyrregular}.
Its 34 datasets span multiple irregularity forms but include related variants and resource-limited failures.
Performance, execution, and statistical dependence must therefore remain distinct.

We introduce the mask-aware state space classifier for irregular time series (MASCIT), a partially selective SSM-based model for naturally irregular TSC.
MASCIT feeds channel-wise observation masks to the selective recurrence and propagates a valid-step mask into multi-head adaptive pooling, aligning encoding and aggregation with the dense representation.
It retains a shallow unidirectional backbone without separate imputation or bidirectional encoding.
Without elapsed-time input, MASCIT models the order and mask geometry of compacted observations rather than physical gaps.
We evaluate predictive and execution performance, capacity ablation, and representation boundaries.

Our contributions are as follows:
\begin{itemize}[leftmargin=*,topsep=2pt,itemsep=1pt,parsep=0pt]
  \item We introduce the first Mamba-based classifier for naturally irregular TSC by conditioning unidirectional selective recurrence on observation masks and excluding invalid steps from multi-head adaptive pooling.
  \item Across 34 datasets and 12 published baselines, MASCIT yielded the strongest aggregate point estimate and was the only evaluated neural model with complete data coverage, establishing Mamba as an effective backbone.
  \item Through regime-wise comparisons and factorial ablations, we show that MASCIT remained first across six irregularity indicators and that partial selectivity was more effective than making all SSM factors input dependent.
\end{itemize}

\section{Related Work}
\label{sec:related-work}

Irregular-series methods differ chiefly in how they encode missingness and time.
Bidirectional recurrent imputation for time series (BRITS) and gated recurrent units with decay (GRU-D) combine recurrence with imputation or decay~\citep{cao2018brits,che2018recurrent}, continuous-time models evolve hidden states between observations~\citep{rubanova2019latent,kidger2020neural}, and attention or graph models learn temporal and inter-sensor relations~\citep{shukla2021multitime,zhang2021raindrop,du2023saits}.
General-purpose transforms, dictionaries, tree ensembles, and temporal convolutions instead rely on representations that need not model irregular time explicitly~\citep{dempster2020rocket,spinnato2024borf,wu2023timesnet}.
PYRREGULAR compares these alternatives under a shared representation and protocol~\citep{spinnato2026pyrregular}.

Mamba conditions selected state space parameters on the current input while retaining linear sequence complexity~\citep{gu2023mamba}.
Our previous work adapted this mechanism to classification through a shallow backbone, gated temporal aggregation, and separate selectivity choices for discretization and input/output projections~\citep{jung2026mambasl}.
MASCIT retains this backbone for dense irregular data.

Benchmark conclusions also depend on dataset composition, preprocessing, and tuning budget~\citep{ruiz2021ueabakeoff}.
For naturally irregular data, resource-limited runs further bias completed-case averages.
Selective SSM comparisons must therefore separate predictive quality, execution coverage, and temporal information exposed by the input~\citep{wen2025interpgn}.

\section{Method}
\label{sec:method}

\subsection{Irregular Time Series Representation}
\label{subsec:problem-representation}

\begin{figure}[!t]
\centering
\includegraphics[width=\columnwidth]{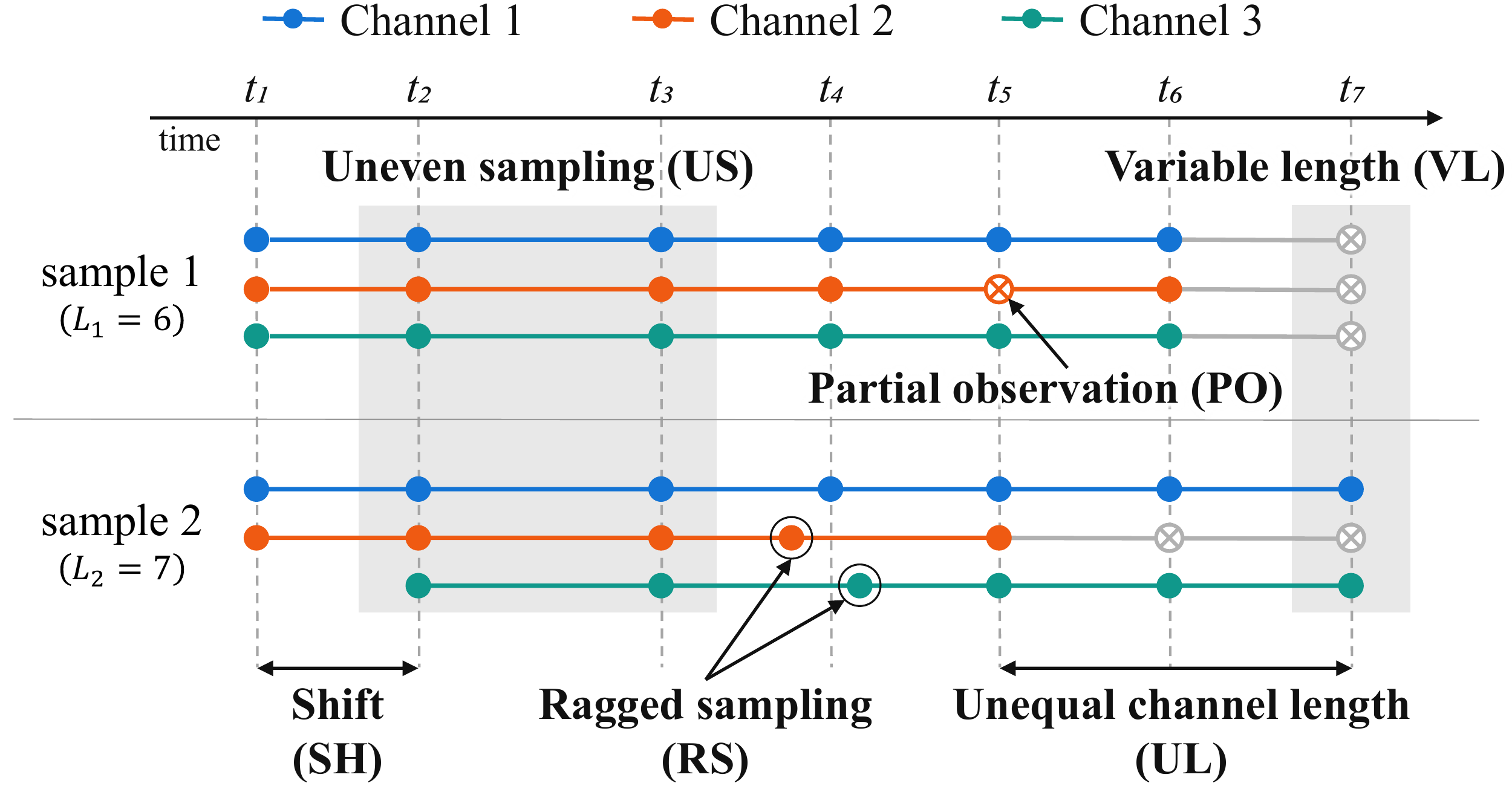}
\caption{Six irregularity indicators used in this study.}
\label{fig:irregularity-types}
\end{figure}

While PYRREGULAR described five irregularity forms, its unequal-length category conflated cross-instance variation with unequal channel lengths within an instance despite defining them distinctly~\citep{spinnato2026pyrregular}.
As shown in Figure~\ref{fig:irregularity-types}, we separate them into variable length and unequal channel length, yielding six overlapping indicators.
Uneven sampling (US) denotes inconstant successive intervals, whereas partial observation (PO) denotes a missing value at an expected timestamp.
Variable length (VL) denotes different compacted lengths across instances, while unequal channel length (UL) denotes different channel observation counts within an instance.
Shift (SH) denotes displaced channel time ranges, and ragged sampling (RS) denotes channel-specific interval sequences.
US, PO, and VL describe irregularity at the instance level, whereas UL, SH, and RS describe mismatched temporal support across channels within a multivariate instance.

Let $\mathcal{D}=\{(\mX_i,\mM_i,y_i)\}_{i=1}^{N}$, where $N$ counts instances, $\mX_i$ is the dense value matrix, $\mM_i\in\{0,1\}^{L\times C}$ is the binary observation mask derived from its observed entries, and $y_i\in\{1,\ldots,K\}$ is the class label.
The adapter constructs $\mX_i$ by sorting the union of timestamps from all $C$ channels and assigning consecutive positions to observed events~\citep{spinnato2026pyrregular}; the model input retains their observedness in $\mM_i$.
Before padding, each matrix has shape $L_i\times C$, where $L_i$ is the compacted length; unobserved values are zeroed and instances are padded to $L=\max_i L_i$.
Compaction preserves values, event order, length, and channel-wise mask geometry but replaces elapsed-time gaps with consecutive indices.

\begin{table*}[!t]
\caption{Dataset taxonomy based on six irregularity indicators for the 34 datasets used in this study. $\mathrm{UL}_0$ denotes the original unequal-length indicator reported by PYRREGULAR and is retained for comparison with the refined taxonomy.}
\label{tab:dataset-taxonomy}
\centering
\resizebox{\textwidth}{!}{%
\addtolength{\tabcolsep}{-3pt}
\begin{tabular}{l*{34}c}
\toprule
\multicolumn{1}{c}{type}
  & \multicolumn{3}{c}{health}
  & \multicolumn{13}{c}{human activity recognition}
  & \multicolumn{9}{c}{mobility}
  & \multicolumn{3}{c}{sensor}
  & \multicolumn{5}{c}{other}
  & \multicolumn{1}{c}{synthetic} \\
\cmidrule(lr){2-4}
\cmidrule(lr){5-17}
\cmidrule(lr){18-26}
\cmidrule(lr){27-29}
\cmidrule(lr){30-34}
\cmidrule(l){35-35}
\multicolumn{1}{c}{dataset}
  & \rotatebox{90}{\texttt{MI3}}
  & \rotatebox{90}{\texttt{P12}}
  & \rotatebox{90}{\texttt{P19}}
  & \rotatebox{90}{\texttt{CT}}
  & \rotatebox{90}{\texttt{GM1}}
  & \rotatebox{90}{\texttt{GM2}}
  & \rotatebox{90}{\texttt{GM3}}
  & \rotatebox{90}{\texttt{GP1}}
  & \rotatebox{90}{\texttt{GP2}}
  & \rotatebox{90}{\texttt{GX}}
  & \rotatebox{90}{\texttt{GY}}
  & \rotatebox{90}{\texttt{GZ}}
  & \rotatebox{90}{\texttt{LPA}}
  & \rotatebox{90}{\texttt{PAM}}
  & \rotatebox{90}{\texttt{PGZ}}
  & \rotatebox{90}{\texttt{SGZ}}
  & \rotatebox{90}{\texttt{AN}}
  & \rotatebox{90}{\texttt{AOC}}
  & \rotatebox{90}{\texttt{APT}}
  & \rotatebox{90}{\texttt{ARC}}
  & \rotatebox{90}{\texttt{GSS\,}}
  & \rotatebox{90}{\texttt{MP}}
  & \rotatebox{90}{\texttt{SE}}
  & \rotatebox{90}{\texttt{TA}}
  & \rotatebox{90}{\texttt{VE}}
  & \rotatebox{90}{\texttt{DD}}
  & \rotatebox{90}{\texttt{DG}}
  & \rotatebox{90}{\texttt{DW}}
  & \rotatebox{90}{\texttt{IW}}
  & \rotatebox{90}{\texttt{JV}}
  & \rotatebox{90}{\texttt{PGE}}
  & \rotatebox{90}{\texttt{PL}}
  & \rotatebox{90}{\texttt{SAD}}
  & \rotatebox{90}{\texttt{ABF}} \\
\midrule
US
  & \cmark & \cmark & \cmark
  & \xmark & \xmark & \xmark & \xmark & \xmark & \xmark & \xmark & \xmark & \xmark & \cmark & \cmark & \xmark & \xmark
  & \cmark & \xmark & \xmark & \xmark & \cmark & \xmark & \cmark & \cmark & \cmark
  & \xmark & \xmark & \xmark
  & \xmark & \xmark & \cmark & \xmark & \xmark
  & \cmark \\
PO
  & \cmark & \cmark & \cmark
  & \xmark & \xmark & \xmark & \xmark & \xmark & \xmark & \xmark & \xmark & \xmark & \xmark & \cmark & \xmark & \xmark
  & \xmark & \xmark & \xmark & \xmark & \xmark & \cmark & \xmark & \xmark & \xmark
  & \cmark & \cmark & \cmark
  & \xmark & \xmark & \xmark & \xmark & \xmark
  & \xmark \\
VL
  & \cmark & \cmark & \cmark
  & \cmark & \cmark & \cmark & \cmark & \cmark & \cmark & \cmark & \cmark & \cmark & \cmark & \cmark & \cmark & \cmark
  & \cmark & \cmark & \cmark & \cmark & \cmark & \xmark & \cmark & \cmark & \cmark
  & \xmark & \xmark & \xmark
  & \cmark & \cmark & \cmark & \cmark & \cmark
  & \xmark \\
UL
  & \cmark & \cmark & \cmark
  & \xmark & \xmark & \xmark & \xmark & \xmark & \xmark & \xmark & \xmark & \xmark & \cmark & \cmark & \xmark & \xmark
  & \xmark & \xmark & \xmark & \xmark & \xmark & \xmark & \xmark & \xmark & \xmark
  & \xmark & \xmark & \xmark
  & \xmark & \xmark & \xmark & \xmark & \xmark
  & \xmark \\
SH
  & \cmark & \cmark & \cmark
  & \xmark & \xmark & \xmark & \xmark & \xmark & \xmark & \xmark & \xmark & \xmark & \cmark & \cmark & \xmark & \xmark
  & \xmark & \xmark & \xmark & \xmark & \cmark & \xmark & \cmark & \cmark & \xmark
  & \xmark & \xmark & \xmark
  & \xmark & \xmark & \cmark & \xmark & \xmark
  & \xmark \\
RS
  & \cmark & \cmark & \cmark
  & \xmark & \xmark & \xmark & \xmark & \xmark & \xmark & \xmark & \xmark & \xmark & \cmark & \cmark & \xmark & \xmark
  & \cmark & \xmark & \xmark & \xmark & \cmark & \xmark & \cmark & \cmark & \cmark
  & \xmark & \xmark & \xmark
  & \xmark & \xmark & \cmark & \xmark & \xmark
  & \xmark \\
\cmidrule(lr){1-35}
$\mathrm{UL}_0$
  & \cmark & \cmark & \cmark
  & \cmark & \cmark & \cmark & \cmark & \cmark & \cmark & \cmark & \cmark & \cmark & \cmark & \cmark & \cmark & \cmark
  & \cmark & \cmark & \cmark & \cmark & \cmark & \cmark & \cmark & \cmark & \cmark
  & \xmark & \xmark & \xmark
  & \cmark & \cmark & \cmark & \cmark & \cmark
  & \xmark \\
\bottomrule
\end{tabular}
\addtolength{\tabcolsep}{+3pt}
}
\end{table*}

\subsection{State Space and Selective State Space Models}
\label{subsec:ssm}

An SSM maps a scalar input $u(t)$ to a state vector $\vh(t)$ of size $d_{\mathrm{state}}$ and a scalar output $o(t)$~\citep{gu2021lssl}:
\begin{equation}
\dot{\vh}(t)=\mA\vh(t)+\mB u(t),\qquad o(t)=\mC\vh(t).
\label{eq:continuous-ssm}
\end{equation}
Here, $\mA\in\mathbb{R}^{d_{\mathrm{state}}\times d_{\mathrm{state}}}$ evolves the state, $\mB\in\mathbb{R}^{d_{\mathrm{state}}}$ writes the input, and $\mC\in\mathbb{R}^{1\times d_{\mathrm{state}}}$ reads the output.
With fixed step size $\Delta$, its discrete form is
\begin{equation}
\vh_{t}=\overline{\mA}\vh_{t-1}+\overline{\mB}u_{t},\qquad o_{t}=\mC\vh_{t},
\label{eq:discrete-ssm}
\end{equation}
where $\overline{\mA}=\mathcal{F}_{A}(\Delta,\mA)$ and $\overline{\mB}=\mathcal{F}_{B}(\Delta,\mA,\mB)$ are discretized parameters.
Fixed parameters make this system time invariant.

For multivariate input, the input projection maps each position to $\vu_{t}\in\mathbb{R}^{d_{\mathrm{model}}}$, where $d_{\mathrm{model}}$ is the projected input dimension.
The componentwise scalar recurrence yields $\vo_t\in\mathbb{R}^{d_{\mathrm{model}}}$.
Mamba linearly projects $\vu_t$ to parameterize $\Delta_t$, $\mB_t$, and $\mC_t$, yielding an input-dependent, time-varying system~\citep{gu2023mamba}.

Our previous work replaced full input dependence with independent time-invariant or time-varying choices for these parameters~\citep{jung2026mambasl}.
For each $P\in\{\Delta,\mB,\mC\}$, let $\phi_P$ be its input-conditioned projection and $\theta_P\in\{0,1\}$ its selectivity switch:
\begin{equation}
P_{t}=(1-\theta_{P})P+\theta_{P}\phi_{P}(\vu_{t}),\qquad P\in\{\Delta,\mB,\mC\}.
\label{eq:partial-selectivity}
\end{equation}
Here, $\theta_P=0$ selects the time-invariant (TI) form and $\theta_P=1$ the input-dependent, time-varying (TV) form.
The three switches define eight patterns; $\Delta$ controls the state-update timescale, while $\mB$ and $\mC$ govern input-to-state and state-to-output routing.
MASCIT retains this factorization and adapts its input and readout to dense irregular series.

\begin{figure}[!t]
\centering
\hspace{+1pt}
\includegraphics[width=\columnwidth]{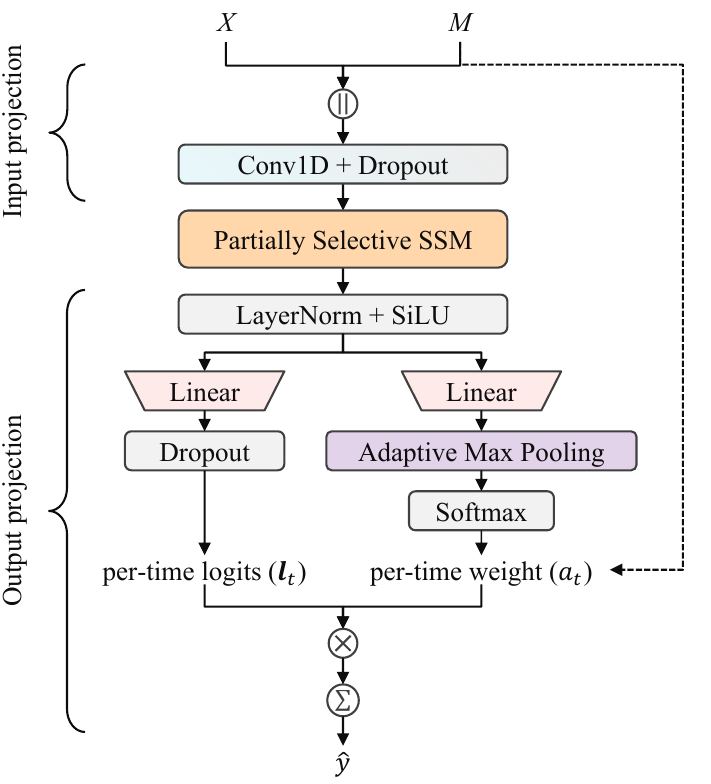}
\caption{Overall structure of MASCIT. MASCIT applies masks to the encoder input and temporal aggregation.}
\label{fig:mascit-architecture}
\end{figure}

\subsection{MASCIT Classifier}
\label{subsec:mascit}

As shown in Figure~\ref{fig:mascit-architecture}, MASCIT adapts our previous classifier to dense irregular series through two mask-aware operations.
First, the input projection embeds $[\mX_i\Vert\mM_i]\in\mathbb{R}^{L\times 2C}$, where $\Vert$ denotes feature-wise concatenation, before the partially selective SSM.
For each instance, $m_t\in\{0,1\}$ indicates whether position $t$ is valid rather than padded.
Second, this valid-step mask excludes padding from gated temporal aggregation.
The length-scaled temporal projection, ordinal positional encoding, single partially selective SSM, and gated readout follow our previous work~\citep{jung2026mambasl}.

The output projection converts each recurrent feature $\vo_{t}$ into class logits $\vl_{t}\in\mathbb{R}^{K}$.
For $H$ gating heads, head $h\in\{1,\ldots,H\}$ has weight $\vw_h\in\mathbb{R}^{d_{\mathrm{model}}}$ and bias $b_h\in\mathbb{R}$.
The largest head score $g_{t}$, normalized temporal weight $\alpha_{t}$, and aggregate logits $\vl$ are
\begin{equation}
\begin{aligned}
g_{t}&=\max_h(\vw_h^\top\vo_{t}+b_h),\\
\alpha_{t}&=\frac{m_{t}\exp(g_{t})}{\sum_r m_r\exp(g_r)}, &
\vl&=\sum_{t}\alpha_{t}\vl_{t}.
\end{aligned}
\label{eq:gated-pooling}
\end{equation}
The denominator spans $r\in\{1,\ldots,L\}$, so $m_r=0$ excludes invalid positions from the readout.
This aggregation ranges from distributed averaging to localized max-like pooling without a length-dependent fully connected readout.
Without timestamp or time-gap input, $\Delta_t$ controls internal state transitions rather than physical elapsed time.

\begin{table*}[!t]
\caption{Aggregate F1 (\%) performance and execution coverage across 34 irregular time series classification datasets.
\textbf{Bold}: best, \underline{underline}: second-best.}
\label{tab:main-results}
\centering
\resizebox{\textwidth}{!}{%
\addtolength{\tabcolsep}{-3pt}
\begin{tabular}{l*{13}{r}}
\toprule
& \multicolumn{6}{c}{non-deep learning} & \multicolumn{7}{c}{deep learning} \\
\cmidrule(lr){2-7}\cmidrule(lr){8-14}
& \multicolumn{2}{c}{distance/kernel} & \multicolumn{1}{c}{tree} & \multicolumn{3}{c}{transform + tree} & \multicolumn{2}{c}{recurrent} & \multicolumn{1}{c}{neural CDE} & \multicolumn{1}{c}{graph} & \multicolumn{1}{c}{transformer} & \multicolumn{1}{c}{convolution} & \multicolumn{1}{c}{SSM} \\
\cmidrule(lr){2-3}\cmidrule(lr){4-4}\cmidrule(lr){5-7}\cmidrule(lr){8-9}\cmidrule(lr){10-10}\cmidrule(lr){11-11}\cmidrule(lr){12-12}\cmidrule(lr){13-13}\cmidrule(l){14-14}
& \multicolumn{1}{c}{\begin{tabular}[c]{@{}c@{}}KNN\\ \citeyearpar{sakoe1978dynamic}\end{tabular}} & \multicolumn{1}{c}{\begin{tabular}[c]{@{}c@{}}SVM\\ \citeyearpar{bagheri2016support}\end{tabular}} & \multicolumn{1}{c}{\begin{tabular}[c]{@{}c@{}}LGBM\\ \citeyearpar{ke2017lightgbm}\end{tabular}} & \multicolumn{1}{c}{\begin{tabular}[c]{@{}c@{}}ROCKET\\ \citeyearpar{dempster2020rocket}\end{tabular}} & \multicolumn{1}{c}{\begin{tabular}[c]{@{}c@{}}BORF\\ \citeyearpar{spinnato2024borf}\end{tabular}} & \multicolumn{1}{c}{\begin{tabular}[c]{@{}c@{}}RIFC\\ \citeyearpar{spinnato2026pyrregular}\end{tabular}} & \multicolumn{1}{c}{\begin{tabular}[c]{@{}c@{}}BRITS\\ \citeyearpar{cao2018brits}\end{tabular}} & \multicolumn{1}{c}{\begin{tabular}[c]{@{}c@{}}GRU-D\\ \citeyearpar{che2018recurrent}\end{tabular}} & \multicolumn{1}{c}{\begin{tabular}[c]{@{}c@{}}NCDE\\ \citeyearpar{kidger2020neural}\end{tabular}} & \multicolumn{1}{c}{\begin{tabular}[c]{@{}c@{}}Raindrop\\ \citeyearpar{zhang2021raindrop}\end{tabular}} & \multicolumn{1}{c}{\begin{tabular}[c]{@{}c@{}}SAITS\\ \citeyearpar{du2023saits}\end{tabular}} & \multicolumn{1}{c}{\begin{tabular}[c]{@{}c@{}}TimesNet\\ \citeyearpar{wu2023timesnet}\end{tabular}} & \multicolumn{1}{c}{\begin{tabular}[c]{@{}c@{}}\modelname\\ (ours)\end{tabular}} \\
\midrule
avg. F1 (\%) & 59.405 & 26.880 & 60.562 & \underline{66.890} & 62.505 & 57.099 & 51.218 & 46.246 & 42.052 & 55.232 & 55.008 & 55.769 & \textbf{69.200} \\
avg. rank & 7.309 & 11.353 & 5.750 & \underline{4.118} & 5.574 & 6.338 & 7.618 & 8.574 & 9.441 & 7.294 & 7.574 & 6.941 & \textbf{3.118} \\
top-1 & 2 & 0 & 4 & \underline{8} & 2 & 3 & 0 & 0 & 2 & 0 & 0 & 0 & \textbf{13} \\
top-3 & 8 & 2 & 9 & \textbf{24} & 13 & 4 & 7 & 3 & 2 & 0 & 4 & 4 & \underline{23} \\
\midrule
data coverage & 29/34 & 31/34 & 34/34 & 34/34 & 34/34 & 34/34 & 32/34 & 32/34 & 32/34 & 32/34 & 32/34 & 32/34 & 34/34 \\
W/L & 24/5 & 29/2 & 24/10 & 19/15 & 25/9 & 26/8 & 29/3 & 29/3 & 28/4 & 28/4 & 27/5 & 28/4 & -- \\
Wilcoxon $p$ & .002 & $<$.001 & .006 & .394 & .026 & .006 & $<$.001 & $<$.001 & $<$.001 & $<$.001 & $<$.001 & $<$.001 & -- \\
\bottomrule
\end{tabular}
\addtolength{\tabcolsep}{+3pt}
}
\end{table*}

\section{Experimental Setup}
\label{sec:experimental-setup}

\subsection{Datasets and Baselines}
\label{subsec:datasets-baseline}

We evaluate all 34 PYRREGULAR datasets spanning healthcare, activity recognition, mobility, sensor, other real-world, and synthetic domains~\citep{spinnato2026pyrregular}.
Table~\ref{tab:dataset-taxonomy} summarizes their six overlapping irregularity indicators, and Appendix~\ref{app:dataset-level-results} gives their full names.
We denote the number of instances, maximum sequence length, and channels by $N$, $L$, and $C$.
We use the official train--test partitions provided by PYRREGULAR.

The 12 baselines span non-deep and deep learning structures.
Non-deep methods comprise $k$-nearest neighbors with dynamic time warping (KNN), a support vector machine with a longest-common-subsequence kernel (SVM), LightGBM (LGBM), bag-of-receptive-fields (BORF), the random interval feature classifier (RIFC), and the random convolutional kernel transform (ROCKET), implemented through MiniRocket~\citep{sakoe1978dynamic,bagheri2016support,ke2017lightgbm,spinnato2024borf,spinnato2026pyrregular,dempster2020rocket,dempster2021minirocket}.
Deep methods comprise BRITS, GRU-D, a neural controlled differential equation (NCDE), Raindrop, self-attention-based imputation for time series (SAITS), and TimesNet~\citep{cao2018brits,che2018recurrent,kidger2020neural,zhang2021raindrop,du2023saits,wu2023timesnet}.
We use benchmark-recommended settings, averaging neural and selected stochastic methods over three runs and evaluating the remainder once~\citep{spinnato2026pyrregular}.

\subsection{Implementation and Evaluation}
\label{subsec:configuration-evaluation}

MASCIT uses one partially selective Mamba layer with fixed $d_{\mathrm{model}}=128$ and $d_{\mathrm{state}}=16$ on every dataset.
For each dataset, we evaluate all eight prespecified TI/TV configurations and report the highest test macro-F1 averaged over three seeds.

Macro-F1 (\%) is the primary metric.
We report data coverage and available-case average F1 (\%); ranks span all 34 datasets, with unavailable runs tied last.
Top-1 and Top-3 count dataset-level placements, while W/L denotes MASCIT wins versus losses over each baseline's available paired datasets.
We apply one-sided Holm-corrected Wilcoxon signed-rank tests for the alternative that MASCIT has a higher F1~\citep{wilcoxon1945test,holm1979simple}.
To reduce cross-dataset scale heterogeneity, Wilcoxon tests use within-dataset z-normalized F1 scores before Holm correction.
Runtime includes fitting and prediction; MASCIT uses the closest available setup to PYRREGULAR.\footnote{PYRREGULAR used an NVIDIA Tesla V100 32 GB GPU; we used an NVIDIA L4 24 GB GPU.}
The source code is available on \href{https://github.com/yoom618/pypots-pyrregular}{Github}.

\section{Results and Discussion}
\label{sec:results-discussion}

\subsection{Performance and Efficiency in Irregular TSC}
\label{subsec:performance-feasibility}

Table~\ref{tab:main-results} summarizes classification performance, execution coverage, and paired comparisons for MASCIT and 12 baselines across 34 PYRREGULAR benchmark datasets.
MASCIT achieves the best average F1 score and rank, as well as the largest number of Top-1 finishes, while ROCKET is the closest published competitor.
The Holm-corrected Wilcoxon tests confirmed statistically significant gains ($p<.05$) over all models except ROCKET.
In addition, MASCIT was the only evaluated neural method with results on all 34 datasets.
Together, these results demonstrate both the inherent capability of Mamba and the effectiveness of the mask-aware selective design for naturally irregular time series classification.

\begin{figure}[!t]
\centering
\includegraphics[width=\columnwidth]{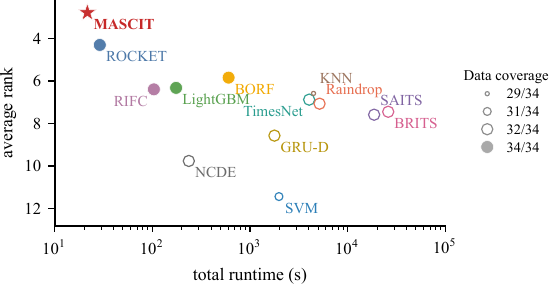}
\caption{Runtime versus average rank on the 29 datasets evaluated by all models; marker area shows data coverage.}
\label{fig:runtime-efficiency}
\end{figure}

To examine whether the predictive gains in Table~\ref{tab:main-results} require additional computational cost, Figure~\ref{fig:runtime-efficiency} relates average rank to mean fit-plus-predict runtime on the 29 datasets successfully evaluated by all models.
MASCIT occupied the most favorable region of the plot, attaining the best average rank at the lowest recorded runtime rather than exchanging efficiency for classification performance.
This pattern supports the efficiency of the shallow selective design, although the mixed hardware prevents a controlled Pareto comparison.

\subsection{Model Structures and Dataset Regimes}
\label{subsec:model-families}

\begin{figure}[!t]
\centering
\includegraphics[width=\columnwidth]{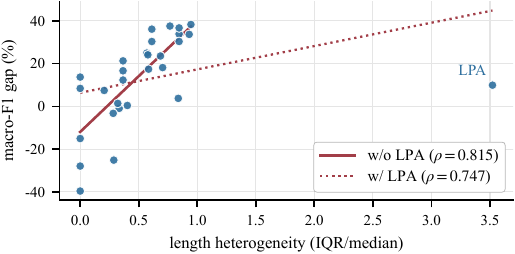}
\caption{Dataset-level relative length dispersion versus the unadjusted generalist--irregular-neural macro-F1 gap (\%p) on 29 common datasets.}
\label{fig:regime-crossover}
\end{figure}

\textbf{Length heterogeneity.} Figure~\ref{fig:regime-crossover} plots the instance-length interquartile range divided by its median against the performance gap between generalist transform/tree pipelines (BORF, LightGBM, RIFC, and ROCKET) and irregularity-oriented neural models (BRITS, GRU-D, NCDE, Raindrop, and SAITS).
The Spearman correlation was positive~\citep{spearman1904}, indicating that greater length heterogeneity is associated with a larger advantage for generalist transform/tree pipelines in this comparison.
The rightmost point, \texttt{LPA}, was a high-leverage person-held-out dataset and visibly flattened the unadjusted line; the result therefore supports a regime association, not a causal benefit of length variation.

\begin{figure}[!t]
\centering
\begin{subfigure}[!t]{\columnwidth}
  \centering
  \includegraphics[width=\linewidth]{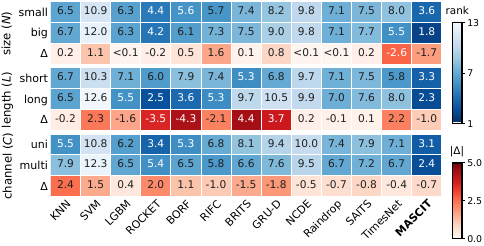}
  \caption{Dataset size, length, and channel count.}
  \label{fig:rank-dataset-properties}
\end{subfigure}
\par\vspace{0.5\baselineskip}
\begin{subfigure}[!t]{\columnwidth}
  \centering
  \includegraphics[width=\linewidth]{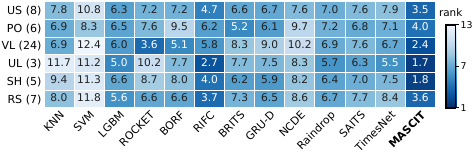}
  \caption{Irregularity indicators.}
  \label{fig:rank-irregularity}
\end{subfigure}
\caption{Average ranks on 29 common datasets; the $\Delta$ in (a) reports the lower-row minus upper-row difference.}
\label{fig:rank-regime-heatmaps}
\end{figure}

\textbf{Dataset scale.} Figure~\ref{fig:rank-regime-heatmaps}(a) compares ranks across splits at $N=500$, $L=360$, and $C=1$.
MASCIT ranked first on both sides and became relatively stronger on larger, multivariate, and longer series, whereas ROCKET shifted most clearly toward longer series.
The ordering is therefore regime dependent even when the aggregate comparison set is held fixed.

\textbf{Irregularity composition.} Figure~\ref{fig:rank-regime-heatmaps}(b) exposes a sharper composition effect.
MASCIT had the lowest point rank within every indicator group, with its strongest values in the sparse \texttt{UL} and \texttt{SH} subsets, whereas ROCKET was most competitive in the dominant \texttt{VL} group and ranked lower elsewhere.
Although ROCKET was the aggregate runner-up in Table~\ref{tab:main-results}, excluding the 17 common-set datasets whose only indicator is \texttt{VL} moved it from second to seventh, while MASCIT remained first.
Thus, ROCKET's aggregate position is composition-sensitive, whereas MASCIT is not confined to the dominant \texttt{VL} stratum.

\begin{table}[!t]
\caption{Model ablation for the two mask interfaces. \textbf{Bold}: best, \underline{underline}: second-best.}
\label{tab:mask-interface-ablation}
\centering
\resizebox{\columnwidth}{!}{%
\addtolength{\tabcolsep}{-2pt}
\begin{tabular}{cccrrrr}
\toprule
model
 & \multicolumn{1}{c}{\begin{tabular}[c]{@{}c@{}}concat\\mask input\end{tabular}}
 & \multicolumn{1}{c}{\begin{tabular}[c]{@{}c@{}}masked\\pooling\end{tabular}}
 & \multicolumn{1}{c}{\begin{tabular}[c]{@{}c@{}}avg.\\F1 (\%)\end{tabular}}
 & \multicolumn{1}{c}{\begin{tabular}[c]{@{}c@{}}avg.\\rank\end{tabular}}
 & \multicolumn{1}{c}{W/L}
 & \multicolumn{1}{c}{\begin{tabular}[c]{@{}c@{}}Wilcoxon\\$p$\end{tabular}} \\
\midrule
\modelname & \cmark & \cmark & \textbf{69.200} & \textbf{1.71} & -- & -- \\
 & \xmark & \cmark & \underline{68.308} & 2.81 & 26/8 & .022 \\
 & \cmark & \xmark & 66.686 & \underline{2.62} & 25/9 & .006 \\
 & \xmark & \xmark & 67.060 & 2.87 & 27/7 & .005 \\
\bottomrule
\end{tabular}
\addtolength{\tabcolsep}{+2pt}
}
\end{table}

\textbf{Mask ablation.} Table~\ref{tab:mask-interface-ablation} shows that MASCIT achieved the best average F1 and rank and outperformed every ablation on most datasets. Removing masked pooling produced the largest decrease in average F1, whereas removing both interfaces did not compound this loss, indicating a non-additive interaction. Nevertheless, all three ablations were inferior under the Holm-corrected Wilcoxon test ($p<.05$), supporting the joint use of mask-conditioned encoding and masked pooling.

\subsection{Sensitivity Analyses}
\label{subsec:additional-analyses}

\begin{figure}[!t]
\centering
\includegraphics[width=\columnwidth]{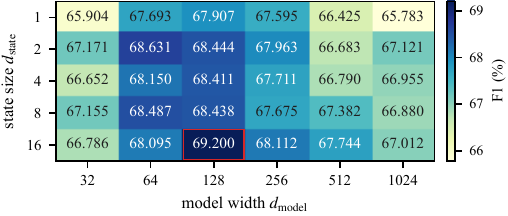}
\caption{Capacity ablation. Each cell value is an average F1 score (\%). Red outlines $128\times16$.}
\label{fig:capacity-response}
\end{figure}

\textbf{Capacity sensitivity.} Figure~\ref{fig:capacity-response} compares $d_{\mathrm{model}}\in\{32,64,128,256,512,1024\}$ and $d_{\mathrm{state}}\in\{1,2,4,8,16\}$ on the 34 datasets.
The high-performing region from widths 64 to 256 remained broad, whereas performance declined at larger widths and changed comparatively little with state size.
Despite the visible color differences, MASCIT remained first in average rank across all $d_{\mathrm{model}}$ and $d_{\mathrm{state}}$ combinations except $(d_{\mathrm{model}},d_{\mathrm{state}})=(32,1)$.
In particular, for each $d_{\mathrm{model}}\in\{64,128,256\}$, the average rank difference between MASCIT and ROCKET exceeded 0.5 across the five $d_{\mathrm{state}}$ values.

\begin{figure}[!t]
\centering
\includegraphics[width=\columnwidth]{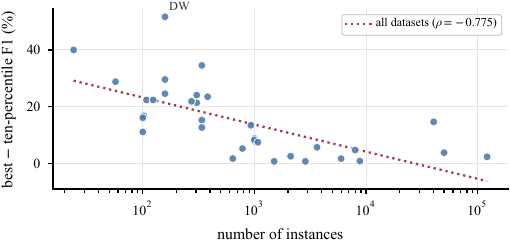}
\caption{Configuration sensitivity versus sample size.}
\label{fig:configuration-sensitivity}
\end{figure}

\begin{table}[!t]
\caption{Noise-adjusted functional ANOVA shares.}
\label{tab:configuration-diagnostics}
\centering
\resizebox{0.8\columnwidth}{!}{%
\begin{tabular}{lrr}
\toprule
component & share & 95\% interval \\
\midrule
$d_{\mathrm{model}}$ & .411 & [.336, .490] \\
$d_{\mathrm{state}}$ & .013 & [.008, .018] \\
time-(in)variance of $\Delta$ & .228 & [.155, .306] \\
time-(in)variance of $B$ & .036 & [.020, .057] \\
time-(in)variance of $C$ & .015 & [.008, .023] \\
\bottomrule
\end{tabular}
}
\end{table}

\begin{figure}[!t]
\centering
\includegraphics[width=\columnwidth]{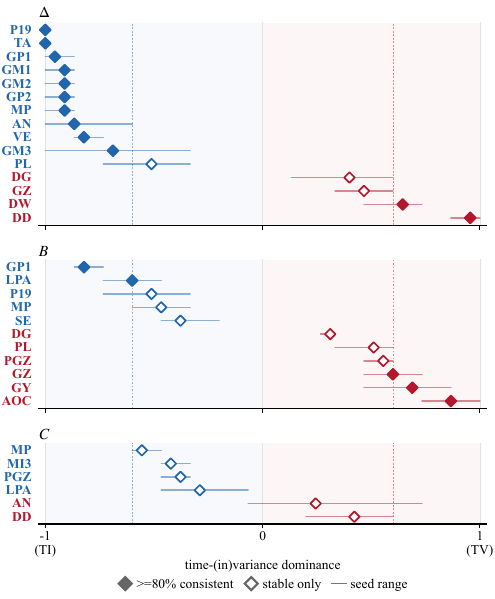}
\caption{Dataset-level TV-minus-TI effects for (top) \(\Delta\), (middle) \(B\), and (bottom) \(C\). Only datasets with consistent directions across runs are shown.}
\label{fig:selectivity-response}
\end{figure}

\textbf{Sample-size sensitivity.} Figure~\ref{fig:configuration-sensitivity} shows that small datasets have a much wider gap between the best and lower-tail configurations, whereas larger datasets are comparatively flat.
This relationship persisted under source exclusions and also appeared in held-seed selection regret.
This indicates selection uncertainty rather than transferable tuning gains.

\textbf{Configuration sensitivity.} Table~\ref{tab:configuration-diagnostics} uses functional analysis of variance (ANOVA) to decompose configuration variation across the complete factorial grids~\citep{hutter2014fanova}.
Main effects accounted for 70.3\% of the noise-adjusted variation, with the remaining 29.7\% attributed to interactions.
Among all individual functional ANOVA components, only $d_{\mathrm{model}}$ (41.1\%), the time-(in)variance of $\Delta$ (22.8\%), and their interaction, $d_{\mathrm{model}}\times\Delta$ (10.5\%), exceeded 5\%.
After capacity, $\Delta$ is therefore the dominant source of configuration variation.

Repeating the two datasets (\texttt{APT} and \texttt{GZ}) with train-only convolution windows preserved the broad capacity and selectivity patterns, indicating that the reported surface was not induced by test-length-dependent window construction.

\subsection{Partial Selectivity}
\label{subsec:capacity-selectivity}

The capacity and configuration sensitivity analyses in Section~\ref{subsec:additional-analyses} identify widths 64--256 as a broad high-performing region and $\Delta$ as the dominant selectivity-related source of variation, motivating a dataset-level analysis of the three selectivity choices within this region.
As shown in Figure~\ref{fig:selectivity-response}, consistent effects are more common and more strongly oriented toward TI for $\Delta$, whereas $B$ splits between TI and TV and $C$ shows no strongly consistent direction.
This dataset-level structure complements the aggregate functional ANOVA result and favors partial over full selectivity.

Global averages conceal local structure.
Within repeated source groups, Gesture-Mid-Air (\texttt{GM1}, \texttt{GM2}, and \texttt{GM3}) and Gesture-Pebble (\texttt{GP1} and \texttt{GP2}) favored TI-$\Delta$, whereas DodgerLoop (\texttt{DD}, \texttt{DG}, and \texttt{DW}) favored TV-$\Delta$.
These contrasts suggest that source and task context shape the preferred form of selectivity beyond the irregularity indicators alone.

As in our previous work, clear dataset-specific TI/TV preferences emerged for both $\Delta$ and $B$, whereas $C$ showed little consistent structure on regularly sampled TSC datasets~\citep{jung2026mambasl}.
However, the distinguishing pattern here is the greater prevalence of TI-$\Delta$ cases despite natural irregularity.
This result suggests that exposing observation irregularity through the mask allows fixed state transitions to remain effective without making $\Delta$ input dependent.

\section{Conclusion}
\label{sec:conclusion}

We proposed MASCIT, a mask-aware state space classifier for irregular time series, which conditions its encoder on channel-wise observation masks and excludes invalid steps from gated aggregation.
MASCIT attained the strongest aggregate point estimate and was the only evaluated neural model with completed results on all 34 datasets.
Factorial analyses favored partial over full selectivity and identified $\Delta$ as the dominant source of selectivity-related variation; together, these findings support a shallow selective state space backbone for dense irregular representations rather than explicit irregular-time modeling.
Nevertheless, the dataset-wise same-test oracle over eight patterns makes the aggregate an optimistic upper envelope, motivating a test-independent selection rule.
\begingroup
\bibliographystyle{apiems-apalike}
\bibliography{apiems-references}
\endgroup

\clearpage
\onecolumn
\appendix
\section{Dataset-Level MASCIT Results}
\label{app:dataset-level-results}

\begin{center}
\captionof{table}{Dataset-level \modelname\,performance and runtime over three seeds. Values are mean $\pm$ sample standard deviation.}
\label{tab:dataset-level-mascit-results}
\centering
\begin{tabular}{lcrrr}
\toprule
dataset & symbol & \multicolumn{1}{c}{macro-F1 (\%)} & \multicolumn{1}{c}{accuracy (\%)} & \multicolumn{1}{c}{runtime (s)} \\
\midrule
MIMIC-III Clinical Database Demo & \texttt{MI3} & $46.387 \pm 19.213$ & $60.606 \pm 10.497$ & $1.9 \pm 0.4$ \\
PhysioNet 2012 & \texttt{P12} & $61.111 \pm 0.939$ & $83.488 \pm 0.504$ & $110.9 \pm 0.5$ \\
PhysioNet 2019 & \texttt{P19} & $68.562 \pm 1.214$ & $92.159 \pm 0.704$ & $863.9 \pm 1.0$ \\
Character-Trajectories & \texttt{CT} & $99.131 \pm 0.335$ & $99.188 \pm 0.314$ & $22.4 \pm 15.4$ \\
Gesture-Mid-Air-D1 & \texttt{GM1} & $63.904 \pm 3.519$ & $64.615 \pm 2.665$ & $7.0 \pm 0.5$ \\
Gesture-Mid-Air-D2 & \texttt{GM2} & $59.743 \pm 2.037$ & $62.564 \pm 1.175$ & $6.7 \pm 0.3$ \\
Gesture-Mid-Air-D3 & \texttt{GM3} & $24.216 \pm 2.663$ & $27.179 \pm 1.776$ & $7.1 \pm 0.5$ \\
Gesture-Pebble-Z1 & \texttt{GP1} & $87.849 \pm 2.540$ & $88.372 \pm 2.664$ & $5.1 \pm 0.5$ \\
Gesture-Pebble-Z2 & \texttt{GP2} & $81.373 \pm 0.944$ & $82.489 \pm 0.731$ & $5.4 \pm 0.5$ \\
All-Gesture-Wiimote-X & \texttt{GX} & $70.390 \pm 1.444$ & $70.286 \pm 1.245$ & $9.1 \pm 2.2$ \\
All-Gesture-Wiimote-Y & \texttt{GY} & $72.356 \pm 2.276$ & $72.571 \pm 2.245$ & $8.4 \pm 1.9$ \\
All-Gesture-Wiimote-Z & \texttt{GZ} & $65.637 \pm 2.525$ & $65.714 \pm 2.199$ & $8.5 \pm 0.8$ \\
Pickup-Gesture-Wiimote-Z & \texttt{PGZ} & $73.828 \pm 4.911$ & $74.667 \pm 4.163$ & $2.4 \pm 0.5$ \\
Shake-Gesture-Wiimote-Z & \texttt{SGZ} & $88.226 \pm 1.005$ & $88.667 \pm 1.155$ & $2.3 \pm 0.5$ \\
Localization Data for Person Activity & \texttt{LPA} & $79.137 \pm 2.981$ & $80.606 \pm 2.777$ & $30.7 \pm 0.4$ \\
PAMAP2 Physical Activity Monitoring & \texttt{PAM} & $56.513 \pm 6.168$ & $68.421 \pm 4.558$ & $8306.8 \pm 1232.9$ \\
Animals & \texttt{AN} & $67.548 \pm 8.214$ & $73.118 \pm 4.928$ & $1.9 \pm 0.7$ \\
Asphalt-Obstacles-Coordinates & \texttt{AOC} & $84.101 \pm 0.842$ & $84.058 \pm 0.822$ & $13.0 \pm 2.1$ \\
Asphalt-Pavement-Type-Coordinates & \texttt{APT} & $96.086 \pm 0.904$ & $96.338 \pm 0.816$ & $100.5 \pm 0.1$ \\
Asphalt-Regularity-Coordinates & \texttt{ARC} & $99.245 \pm 0.077$ & $99.245 \pm 0.077$ & $102.5 \pm 36.4$ \\
GeoLife Supervised & \texttt{GSS} & $11.094 \pm 0.473$ & $42.735 \pm 0.536$ & $11512.3 \pm 8371.3$ \\
Melbourne Pedestrian & \texttt{MP} & $92.683 \pm 0.515$ & $92.770 \pm 0.519$ & $32.6 \pm 0.9$ \\
GPS Data of Seabirds & \texttt{SE} & $61.327 \pm 0.528$ & $73.737 \pm 1.750$ & $16.7 \pm 0.5$ \\
Taxi & \texttt{TA} & $24.521 \pm 0.452$ & $51.801 \pm 0.004$ & $799.6 \pm 325.6$ \\
Vehicles & \texttt{VE} & $61.542 \pm 2.011$ & $76.232 \pm 0.502$ & $7.0 \pm 3.6$ \\
Dodger-Loop-Day & \texttt{DD} & $44.746 \pm 7.296$ & $47.083 \pm 6.884$ & $3.3 \pm 0.4$ \\
Dodger-Loop-Game & \texttt{DG} & $61.935 \pm 25.885$ & $67.150 \pm 17.123$ & $1.1 \pm 0.8$ \\
Dodger-Loop-Weekend & \texttt{DW} & $95.476 \pm 2.651$ & $96.377 \pm 2.174$ & $1.2 \pm 0.7$ \\
Insect-Wingbeat & \texttt{IW} & $65.525 \pm 0.285$ & $65.633 \pm 0.240$ & $898.2 \pm 3.9$ \\
Japanese-Vowels & \texttt{JV} & $97.650 \pm 0.397$ & $97.748 \pm 0.312$ & $6.1 \pm 2.3$ \\
Productivity Prediction of Garment Employees & \texttt{PGE} & $100.000 \pm 0.000$ & $100.000 \pm 0.000$ & $0.8 \pm 0.4$ \\
PLAID & \texttt{PL} & $59.478 \pm 1.574$ & $63.253 \pm 0.919$ & $27.0 \pm 4.0$ \\
Spoken-Arabic-Digits & \texttt{SAD} & $99.077 \pm 0.264$ & $99.075 \pm 0.266$ & $83.7 \pm 22.8$ \\
Alembics Bowls Flasks & \texttt{ABF} & $32.390 \pm 1.711$ & $33.630 \pm 2.478$ & $1.8 \pm 1.0$ \\
\bottomrule
\end{tabular}%
\end{center}

\end{document}